\documentclass{article}
\usepackage{ijcai19}

\usepackage{times}
\usepackage{soul}
\usepackage{url}
\usepackage[hidelinks]{hyperref}
\usepackage[utf8]{inputenc}
\usepackage[small]{caption}
\usepackage{graphicx}
\usepackage{amsmath}
\usepackage{amssymb}
\usepackage{amsthm}
\usepackage{amsopn}
\usepackage{booktabs}
\usepackage{algorithm}
\usepackage{algorithmic}
\usepackage{enumerate}
\usepackage{color}
\usepackage{multirow}
\usepackage{hyperref}
\usepackage{diagbox} 
\newcommand{\eg}{\emph{e.g. }}

\newcommand{\ie}{\emph{i.e. }}

\newcommand{\wrt}{w.r.t. }

\title{Attribute Aware Pooling for Pedestrian Attribute Recognition}

\author{
Kai Han$^1$, Yunhe Wang$^1$, Han Shu$^1$, Chuanjian Liu$^1$, Chunjing Xu$^1$, Chang Xu$^2$\footnote{We thank anonymous reviewers for their helpful comments. Chang Xu was supported by the Australian Research Council under Project DE180101438.}
\affiliations
$^1$Huawei Noah's Ark Lab\\
$^2$School of Computer Science, FEIT, University of Sydney, Australia\\
\emails
\{kai.han, yunhe.wang, han.shu, liuchuanjian, xuchunjing\}@huawei.com,
c.xu@sydney.edu.au
}

\begin{document}

\maketitle

\begin{abstract}
This paper expands the strength of deep convolutional neural networks (CNNs) to the pedestrian attribute recognition problem by devising a novel attribute aware pooling algorithm. Existing vanilla CNNs cannot be straightforwardly applied to handle multi-attribute data because of the larger label space as well as the attribute entanglement and correlations. We tackle these challenges that hampers the development of CNNs for multi-attribute classification by fully exploiting the correlation between different attributes. The multi-branch architecture is adopted for fucusing on attributes at different regions. Besides the prediction based on each branch itself, context information of each branch are employed for decision as well. The attribute aware pooling is developed to integrate both kinds of information. Therefore, attributes which are indistinct or tangled with others can be accurately recognized by exploiting the context information. Experiments on benchmark datasets demonstrate that the proposed pooling method appropriately explores and exploits the correlations between attributes for the pedestrian attribute recognition.
\end{abstract}

\section{Introduction}
Pedestrian attribute recognition has appealed much research effort due to its continuing demands for intelligent video surveillance~\cite{layne2012person,peng2016joint,liu2017hydraplus}. It is a challenging task because of the large variance in pedestrian images, such as the viewpoint, lighting or pose changes. Thanks to the development of deep learning, pedestrian attribute recognition based on convolutional neural networks (CNNs) has made tremendous progress on the benchmark datasets, \eg~PETA~\cite{deng2014pedestrian}, RAP~\cite{li2016richly}, and PA-100K~\cite{liu2017hydraplus}. However, existing methods still have a long way to the practical application where the scenario is pretty complex.


\begin{figure*}[t]
\centering
\includegraphics[width=0.85\linewidth]{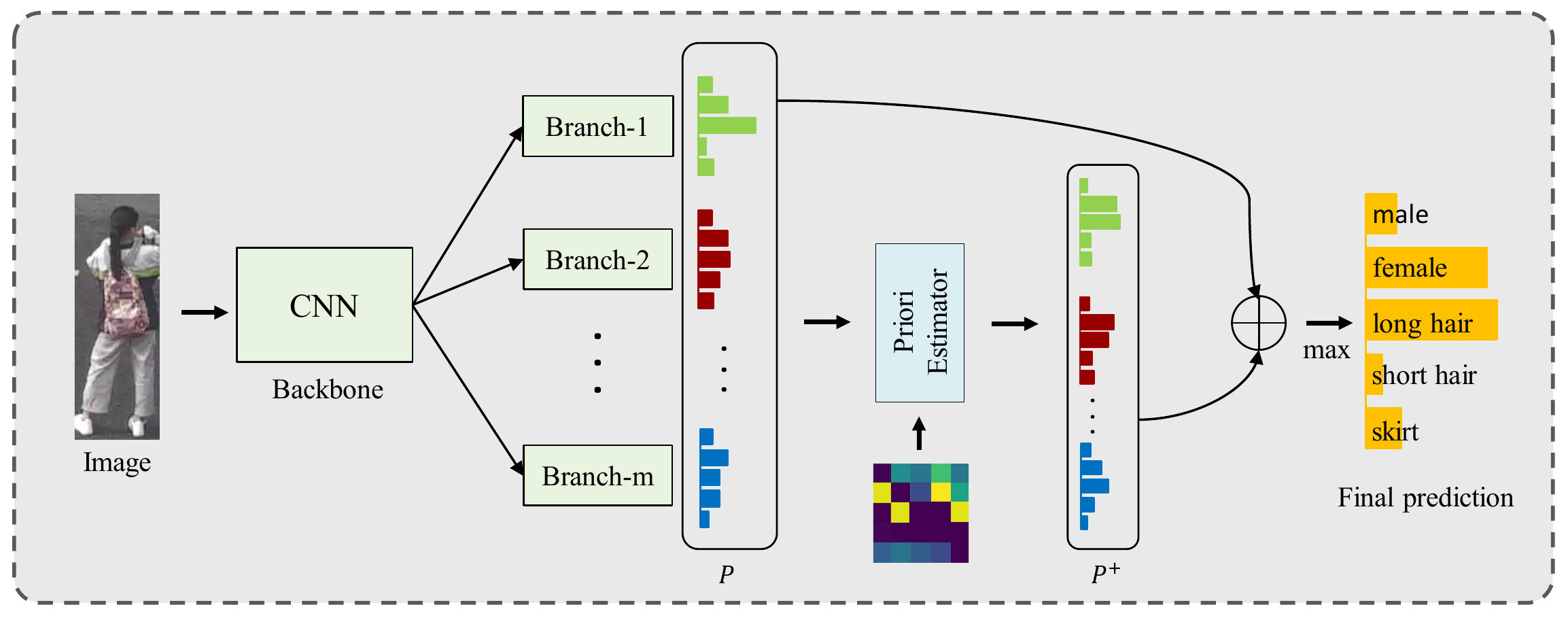}
\vspace{-0.em}
\caption{The diagram of the proposed attribute aware pooling approach. The input instance is fed into a shared CNN and produce multiple predictions with multi-branch architecutre (details in Fig.~\ref{Fig:multi-branch}). Afterward, we exploit the co-occurrence priori to integrate these conditional probabilities, thereby producing a attribute aware pooling estimation.}
\label{Fig:intro}
\vspace{-0.3cm}
\end{figure*}

Extraordinary capability of CNN to accomplish the single-label image classification task has been comprehensively validated~\cite{ResNet,fasterrcnn,wang2016cnnpack}. An example in the ImageNet Dataset~\cite{ImageNet} has only one label, such as ``dog", ``cat" or ``car". However, pedestrian often contains multiple attributes, \eg ``age", ``gender'' and ``clothing". Moreover, the fact that these attributes usually do not correspond to some certain objects will bring in hidden semantic recognition problem. Multi-label pedestrian attribute classification is thus a more practical and challenging problem to be thoroughly investigated. The superiority of CNNs in the standard single-label natural image classification cannot be straightforwardly extended to the attribute classification scenario due to the difference in input image domains and problem settings. Specifically, in multi-attribute recognition, candidate attributes may locate in different regions of the given image and the label space of a $k$-category classification task becomes $2^k$, which requires more training data and parameters for establishing a deep neural network to achieve a comparable accuracy. Current CNN solutions for the pedestrian attribute classification problem mainly focus on extracting high-level features from the entire image~\cite{sudowe2015person,li2015multi,xiao2016learning}, and multi-branch architecture for different attribute groups~\cite{zhu2015multi,lu2017fully,zhao2018grouping,han2018attribute}, to accomplish the subsequent classification.

The multi-branch manner for pedestrian attribute recognition achieves better performance via sharing the bottom $l$ layers then breakuping into tree-like architecture, resulting in task-specific sub-networks for similar attribute groups. This is inspired by the common observation that bottom layers in CNNs mainly extract low level visual features which can be shared across different tasks, while the top layers capture high level semantic features that are more task specific.~\cite{hand2017attributes} proposed a multi-task deep convolutional neural network (MCNN) for facial attribute classification.~\cite{lu2017fully} dynamically created a tree-like deep architecture where similar tasks reside in the same branch until the top layers.~\cite{Wang_2017_ICCV,zhao2018grouping} utilized a CNN-RNN model to take advantage of the intragroup mutual exclusion and inter-group correlation. These methods mine the correlations of attributes, but ignore the prior knowledge underlying the attribute data.

In this paper, we propose to develop a new CNN architecture for pedestrian multi-attribute recognition by exploring the correlation between different attributes, \ie attribute co-occurrence priori~\cite{co1,co2,co3,co4}. In particular, there is a high probability that some attributes that often co-occur in the training set will also co-occur in the testing set. {\color{black}For example, given a pedestrian annotated with ``long hair", ``skirt" and ``high-heeled shoes", we can easily deduce its gender attribute as ``female''.} Based on this insightful observation, we develop a novel attribute aware pooling method (AAP) for integrating the information from different predictions, namely CoCNN. More specifically, the base network follows the multi-branch architecture, and then the context information in these branch is collected to estimate the attribute probabilities, which is then combined with the individual estimations of each branch for improving the resulting decision. The diagram of the proposed scheme is shown in Fig.~\ref{Fig:intro}. Experimental results on benchmarks demonstrate the superiority of the proposed algorithm over the state-of-the-art methods for pedestrian attribute classification.

\section{Preliminaries}
We first briefly introduce some related works on CNN and then build the relationship between the output of a conventional CNN and the proposed attribute aware pooling method.

Let $\mathcal{X}$ and $\mathcal{Y}$ denote instance space and $k$-label space, respectively. Given a labeled training set with $n$ instances, $\{(\mathbf{x}^1,\mathbf{y}^1),(\mathbf{x}^2,\mathbf{y}^2),...,(\mathbf{x}^n,\mathbf{y}^n)\}$, where $\mathbf{x}^i$ is the $i$-th instance and $\mathbf{y}^i$ denotes its label vector which is a $k$-dimensional binary vector, \ie $\mathbf{y}^i = [\mathbf{y}_1^i,...,\mathbf{y}_k^i]$ and $\mathbf{y}_k^i \in \{0, 1\}$. Denote the feature of $\mathbf{x}_i$ calculated by a given CNN $\mathcal{N}$ as $x^i = \mathcal{N}(\mathbf{x}^i) \in \mathbb{R}^{d}$, the conventional probability hypothesis for the $j$-th attribute is
\begin{equation}
\hat{\mathbf{y}}^i_j = \Pr(\mathbf{a}^j |x^i) = \frac{1}{1+e^{-\theta_j^Tx^i}},
\end{equation}
where $\mathbf{a}^j$ is the $j$-th attribute in the dataset corresponding to the $j$-th dimensionality of the label space $\mathcal{Y}$, $\theta_j\in\mathbb{R}^{d}$ is the parameter of linear classification layer. 

In practice, for a given image $\mathbf{x}$ (image index $i$ is omitted to simplify notation), it is difficult to accurately estimate its conditional probabilities \wrt all attributes labels simultaneously, due to the tangle between attributes and the deformation of objects, \eg~occlusions, illumination changes, rotations, and scale transforms. To address this problem, the multi-branch architecture is utilized in this work, \ie~we have $m$ branches $\{{b}_1,...,{b}_m\}$. The lower layers are shared across branches and the top layers are separated to focus on the attributes at different positions, as shown in Fig.~\ref{Fig:intro}. The detailed multi-branch architecture is given in the experiments section. Formally for each branch, we can construct the following $m\times k$ matrix, 
\begin{equation}
P = \left[
\begin{array}{cccc}     
\Pr(\mathbf{a}^1 |b_1)& \Pr(\mathbf{a}^2 |b_1) & \hdots &\Pr(\mathbf{a}^k |b_1) \\
\Pr(\mathbf{a}^1 |b_2)& \Pr(\mathbf{a}^2 |b_2)& \ddots   &\vdots\\
\vdots&   \ddots  & \ddots &   \\
\Pr(\mathbf{a}^1 |b_m)& \hdots&  & \Pr(\mathbf{a}^k |b_m)  \\
\end{array}\right],
\label{Eq:P}
\end{equation}
where $\Pr(\mathbf{a}^j |b_l)$ stands for the conditional probability of $\mathbf{a}^j$ given the feature of $l$-th branch, and each row in $P$ corresponds to the softmax output of the branch $b_l$. Moreover, conditional probabilities of the whole image $\mathbf{x}$ \wrt different attributes can be estimated through a max-pooling operation: 
\begin{equation}
\Pr(\mathbf{a}^j |\mathbf{x}) \approx \max_{l\in\left[1,m\right]} \Pr(\mathbf{a}^j |b_l) = \max_{l\in\left[1,m\right]} P_{l,j}.
\label{Eq:maxp}
\end{equation}

It is instructive to note that the accuracy of Eq.~\ref{Eq:maxp} strongly depends on the prediction result of different branches. These branches are extracted from input images by exploiting empirical knowledge such as position information. However, attributes tangled with each other are difficult to be separated by branches accurately.  Hence, the estimated conditional probability vector using Eq.~\ref{Eq:maxp} could be biased when the $j$-th attribute in $\mathbf{x}$ has not be well separated. Elements in the $j$-th column of $P$ will be extremely small, which leads to the $\Pr(\mathbf{a}^j |\mathbf{x})$ after the max-pooling being small as well.

A widely accepted fact in the filed of multi-label classification~\cite{co4,wang2013visual,co3} is that correlated labels have a strong probability to simultaneously occur in real-world images. {\color{black}In practice, for an attribute $\mathbf{a}^j$, we can use the other attributes to estimate the conditional probability \wrt the $j$-th attribute. The auxiliary estimation is generated as follow: 
\begin{equation}
P_{l,j}^+ = \frac{1}{s} \sum_{i=1}^kS_l^i\Pr(\mathbf{a}^j|\mathbf{a}^i),
\label{Eq:p+}
\end{equation}
where $S_l^i\in\{0,1\}$ is an indicator and $S_l^i=1$ denotes $\mathbf{a}^i$ appears in the image, and vice versa, and $s = \sum_{i=1}^kS_l^i$ is the number of the positive attributes. $\Pr(\mathbf{a}^j | \mathbf{a}^i)$ is the conditional probability of two attributes or labels which can be pre-calculated by accounting the number of their co-occurrence in the dataset. $P^+$ is an $m\times k$ matrix, which can be applied for refining the prediction result of conventional CNNs for the multi-label pedestrian attribute classification problem. Since the co-occurance information $S_l^j$ of any branch $b_l$ is not given in the dataset, Eq.~\ref{Eq:p+} is hard to calculate directly.} We therefore propose a novel attribute aware pooling method which first integrates the co-occurance information of every branch $\{b_{1},...,b_{m}\}$ and the priori knowledge, and then generates the estimation of the whole image by exploiting the co-occurrence priori over different attributes.

\section{Attribute Aware Pooling}
To further explore the correlation between attributes using the priori knowledge in the real data, in this section, we first build conditional probability matrices \wrt different attributes using the co-occurrence priori and then develop the new attribute aware pooling method.

\subsection{Co-occurrence Priori Embedding}
\begin{figure}[t]
	\centering
	\includegraphics[width=0.8\linewidth]{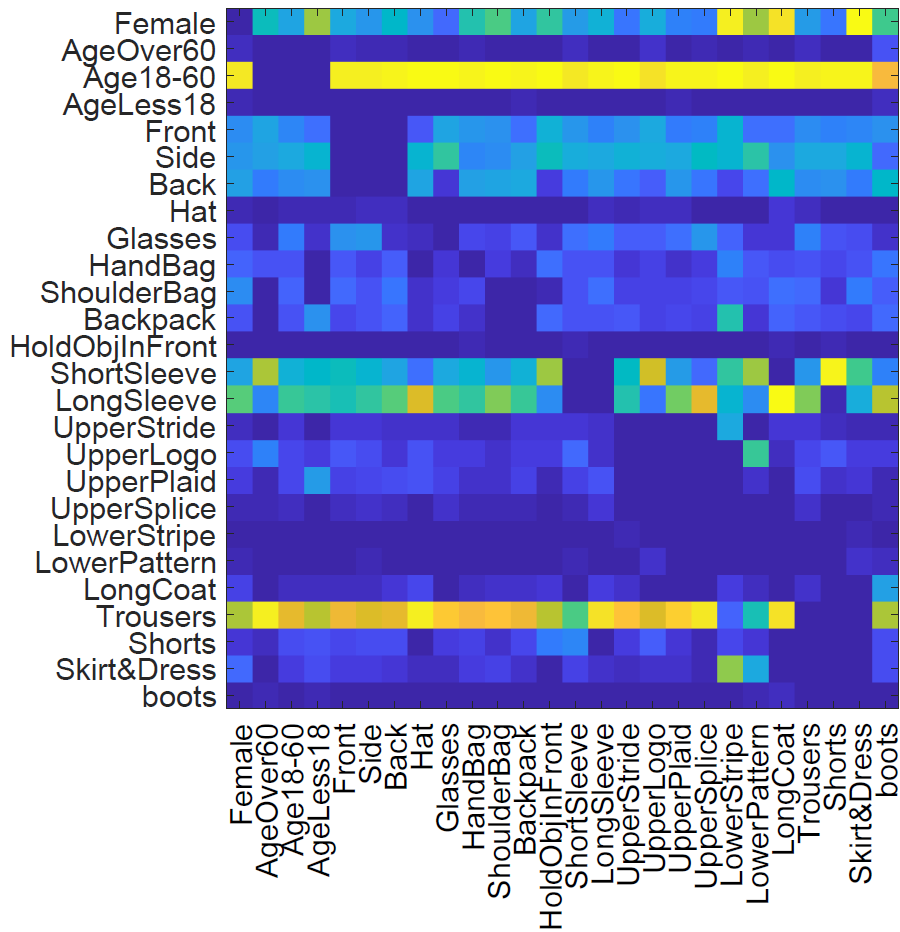}
	\caption{ $C$ learned on PA-100K dataset. Darker color means smaller value.}
	\label{Fig:C}
	\vspace{-0.5cm}
\end{figure}

Given a training set with $k$ labels and $n$ instances, let $N_i$ denote the number of the $i$-th label occur in the dataset, and $p_i = \Pr(\mathbf{a}^i)=N_i/n$ be the probability of the $i$-th label. By denoting the co-occurrence number of the $i$-th label and the $j$-th label in the dataset as $N_{i,j}$, we can construct a matrix $J\in\mathbb{R}^{k\times k}$, where $J_{i,j} =\Pr(\mathbf{a}^i,\mathbf{a}^j)= N_{i,j}/n$ is the joint probability of $\mathbf{a}^i$ and $\mathbf{a}^j$. Subsequently, we can obtain the conditional probability of an attribute given another attribute as
\begin{equation}
C_{i,j} = \Pr(\mathbf{a}^i|\mathbf{a}^j) = \frac{\Pr(\mathbf{a}^i, \mathbf{a}^j)}{\Pr(\mathbf{a}^j)} = \frac{J_{ij}}{p_j}.
\end{equation}
After constructing this matrix, we can further estimate the auxiliary probability $P^+_{l,j}$ by exploiting the context information of the $l$-th branch $\mathbf{x}_l$. Fig.~\ref{Fig:C} shows the matrix $C$ discovered on PA-100K dataset~\cite{liu2017hydraplus}, we can see that some attributes frequently co-occur in the dataset, \eg ``Female'' and ``Skirt\&Dress'', ``LongSleeve'' and ``Trousers''.

In the $k$-label classification task, given an instance with $m$ branches $\{b_1,...,b_m\}$, the softmax classifier outputs $m$ $k$-dimensional vectors, which are stacked into an $m\times k$ matrix $P$ as shown in Eq.~\ref{Eq:P}. {\color{black}For any branch $b_l$, there are $m-1$ branches in $\mathbf{x}$ that surround with $b_l$ and we denote these branches as $\tilde{b}_l$. The probability of the occurrence of attribute $\mathbf{a}^j$ in $\tilde{b}_l$ can be calculated by
\begin{equation}
\Pr(\mathbf{a}^j_l)=\Pr(\mathbf{a}^j|b_1,...,b_{l-1},b_{l+1},...,b_{m}).
\end{equation}
However, this high-order posterior probability cannot be accurately calculated. Alternatively, we use the following locally max-pooling as an approximation: 
\begin{equation}
Q_{l,j} = \Pr(\mathbf{a}^j_l) \approx \max_{i\neq l} \Pr(\mathbf{a}^j|b_i).
\label{Eq:localmax}
\end{equation}
Attributes identification in $\tilde{b}_l$ can be easily accomplished through the following threshold function with the parameter $\tau$:
\begin{equation}
S_l^j  = \bigg\{
\begin{array}{l}
1, \quad \text{if} ~~ Q_{l,j} >\tau,\\
0, \quad \text{otherwise}.\\
\end{array}
\end{equation}
and we can use this indicator matrix for calculating Eq.~\ref{Eq:p+}. Compared with the output conditional probability $P_{l,j}$ of the $l$-th branch $b_l$ given the $j$-th attribute derived from CNNs, $P_{l,j}^+$ is calculated by exploiting outputs of other branches, and can be taken as another important auxiliary predictor. 

However, given the biased recognition of the branches, this hard threshold strategy may bring in inaccurate classification result and the introduction of the hyper-parameter $\tau$ will make $s$ various for different branches. In addition, since attributes have some conflicts, absence of an attribute may lead to the appearance of another one. Therefore, we use the total probability theorem to modify $P_{l,j}^+$ as follow:
\begin{equation}
\begin{aligned}
P_{l,j}^+ = \frac{1}{k}\sum_{i=1}^k&\Pr(\mathbf{a}^i_l)\Pr(\mathbf{a}^j|\mathbf{a}^i)\\&+\frac{1}{k}\sum_{i=1}^k\Pr(\tilde{\mathbf{a}}^i_l)\Pr(\mathbf{a}^j|\tilde{\mathbf{a}}^i)
\label{Eq:total}
\end{aligned}
\end{equation}
where $\tilde{\mathbf{a}}^i$ indicates that $\mathbf{a}^i$ does not occur, and the second term can be calculated as
\begin{equation}
\begin{aligned}
\Pr(\tilde{\mathbf{a}}^i_l)&\Pr(\mathbf{a}^j|\tilde{\mathbf{a}}^i) = \Pr(\tilde{\mathbf{a}}^i_l)\frac{\Pr(\mathbf{a}^j,\tilde{\mathbf{a}}^i)}{\Pr(\tilde{\mathbf{a}}^i)}\\
&=\Pr(\tilde{\mathbf{a}}^i_l)\frac{\Pr(\mathbf{a}^j)-\Pr(\mathbf{a}^j,\mathbf{a}^i)}{1-\Pr(\mathbf{a}^i)}\\
& = (1-\Pr(\mathbf{a}^i_l))\frac{p_j-J_{i,j}}{1-p_i},
\end{aligned}
\end{equation}
and we can first calculate the matrix 
\begin{equation}
\tilde{C}_{i,j} =\Pr(\mathbf{a}^i|\tilde{\mathbf{a}}^j)=\frac{p_i-J_{i,j}}{1-p_j}
\end{equation}
based on the training set in advance for convenience.
}
By exploiting the pre-defined two matrices $C$ and $\tilde{C}$, the additional estimated probability matrix $P^+$ can be simply calculated as
\begin{equation}
P^+ = \frac{1}{k}\left[QC+(\mathbf{1}-Q)\tilde{C}\right],
\label{Eq:P+}
\end{equation}
where $\mathbf{1}$ is a $k\times k$ full one matrix.

Eq.~\ref{Eq:P+} is estimated by exploiting context information and co-occurrence tables, which provides another approach to estimate the classification result of the input instance $\mathbf{x}$. The attribute aware pooling method is thus developed to combine the two decision matrices of different properties. Formally, the output of the attribute aware pooling layer is
\begin{equation}
\hat{P} = P+\lambda P^+,
\label{Eq:hatP}
\end{equation}
where $\lambda$ is the parameter for balancing the prediction results of the branch itself and context information from the other branches. We will further test its impact on the classification accuracy in the following section experimentally.

Finally, we assemble predictions of every branch to form a $k$-dimensional vector. Since every element in $\hat{P}$ is greater than zero, for a given image $\mathbf{x}^i$, the output of the proposed method is 
\begin{equation}
\begin{aligned}
\hat{\mathbf{p}}_j^i = \Pr(\mathbf{a}^j|\mathbf{x}^i)=\frac{\max_{l\in[1,m]}\hat{P}_{l,j}}{\sum_{j=1}^k\max_{l\in[1,m]}\hat{P}_{l,j}},\\
\end{aligned}
\label{Eq:finalest}
\end{equation}
where $M\in\{0,1\}^{m\times k}$ is the max-pooling projection which has only one 1 per column. Let $\mathcal{E}^i = \text{diag}(\hat{P}^TM)$, and we have
\begin{equation}
\hat{\mathbf{p}}^i=\frac{\mathcal{E}^i}{||\mathcal{E}^i||_1},
\vspace{1em}
\end{equation}
where $||\cdot||_1$ is the conventional $\ell_1$ norm for vectors. The ground-truth conditional probabilities of $\mathbf{x}^i$ is $\mathbf{p}^i=\mathbf{y}^i/||\mathbf{y}^i||_1$. Therefore, the loss function of the network using the proposed method is
\begin{equation}
\begin{aligned}
J =& \frac{1}{2n}\sum_{i=1}^n\sum_{j=1}^k\left(\hat{\mathbf{p}}_j^i -\mathbf{p}_j^i \right)^2 \\=& \frac{1}{2n}\sum_{i=1}^n||\hat{\mathbf{p}}^i-\mathbf{p}^i||_2^2.
\end{aligned}
\label{Eq:finalloss}
\end{equation}

By optimizing the above loss function using the mini-batch SGD approach, we can fine-tune filters in the neural network $\mathcal{N}$ and obtain a new CNN with improvement in attribute recognition accuracy. The diagram of the proposed attribute aware pooling architecture is shown in Fig.~\ref{Fig:intro}.

\subsection{Back Propagation}

We proceed to introduce the optimization method for CNNs with the proposed attribute aware pooling method. Stochastic gradient descent (SGD) needs to calculate the gradient of the loss function and then utilizes the back propagation strategy for updating parameters. We next detail the gradient and the back propagation of the proposed method.

In fact, the proposed attribute aware pooling method is attached after the softmax layer of a general CNN on $m$ branches and followed by some pooling and probability calculations as shown in Fig.~\ref{Fig:intro}. Therefore, we do not have to modify any additional convolution layer and filters, and we only need to calculate gradient of the loss function in the last layer \wrt the individual branch estimation matrix $P$ in Eq.~\ref{Eq:P}. Commonly, this gradient can be computed using the standard chair rule, \ie
\begin{equation}
\frac{\partial J}{\partial P}=\frac{\partial J}{\partial \hat{\mathbf{p}}}\cdot \frac{\partial \hat{\mathbf{p}}}{\partial \mathcal{E}} \cdot\frac{\partial\mathcal{E}}{\partial\hat{P}} \cdot \frac{\partial \hat{P}}{\partial P}.
\end{equation}
We first calculate the gradient of the first two terms for a given instance $\mathbf{x}$ as follow:
\begin{equation}
\frac{\partial J}{\partial \mathcal{E}} = \frac{\hat{\mathbf{p}}-\mathbf{p}}{||\mathcal{E}||_1}-\frac{(\hat{\mathbf{p}}-\mathbf{p})^T\mathcal{E}}{||\mathcal{E}||_1^2},
\end{equation}
where $||\mathcal{E}||_1 = \sum_{j=1}^k |\mathcal{E}_j|$, and $\mathcal{E}\in\mathbb{R}^{k\times1}$. Subsequently, the gradient of $\hat{P}$ is calculated as follow: 
\begin{equation}
\frac{\partial J}{\partial \hat{P}} = M\;\text{diag}\left(\frac{\hat{\mathbf{p}}-\mathbf{p}}{||\mathcal{E}||_1}-\frac{(\hat{\mathbf{p}}-\mathbf{p})^T\mathcal{E}}{||\mathcal{E}||_1^2}\right),
\end{equation}
where the variables in $\text{diag}(\cdot)$ make up a $k$-dimensional vector and $M$ is the max-pooling projection which maps the gradient into a $m\times k$ sparse matrix. 

As for the gradient of $P$, consider $Q_l = \text{diag}(P^T S_l)$ where $S_l\in\{0,1\}^{m\times k}$ is the $l$-th mask \wrt the locally max-pooling (Eq.~\ref{Eq:localmax}) for generating the $l$-th row of $Q$. Elements in the $l$-th row of $S_l$ are zeros. The $l$-th row of $\hat{P}$ can be reformulated as
\begin{equation}
\begin{aligned}
\hat{P}_l &= P_l+\lambda P_l^+ = P_l+\lambda (Q_lC+(1-Q_l)\tilde{C})\\
 = &P_l+\frac{\lambda}{k} \left[ \text{diag}(P^TS_l)C+(\mathbf{1}-\text{diag}(P^TS_l))\tilde{C}\right].
\end{aligned}
\end{equation}
Removing terms irrelevant to $P$, the $j$-th element in $P_l^+$ can be simplified as
\begin{equation}
\begin{aligned}
P_{l,j}^+ &=\text{diag}(P^TS_l)C_j-\text{diag}(P^TS_l)\tilde{C}_j\\
&= \text{Tr}(P^TS_l\;\text{diag}(C_j))-\text{Tr}(P^TS_l\;\text{diag}(\tilde{C}_j)),
\end{aligned}
\end{equation}
where $C_j$ and $\tilde{C}_j$ are the $j$-th columns in $C$ and $\tilde{C}$, respectively. The gradient of $P_{l,j}^+$ to $P$ is
\begin{equation}
\begin{aligned}
\frac{\partial P_{l,j}^+}{\partial P} &= S_l\;\text{diag}(C_j)-S_l\;\text{diag}(\tilde{C}_j)\\
&=S_l\;\text{diag}(C_j-\tilde{C}_j).
\end{aligned}
\end{equation}
Therefore, the gradient of the classification loss \wrt $P$ is 
\begin{equation}\small
\frac{\partial J}{\partial P}= \frac{\partial J}{\partial \hat{P}}+\frac{\lambda}{k}\sum_{l=1}^{m}\sum_{j=1}^{k}\frac{\partial J}{\partial \hat{P}_{l,j}}S_l\;\text{diag}(C_j-\tilde{C}_j),
\label{Eq:gradP}
\vspace{-0.5em}
\end{equation}
and $P$ can be updated as
\begin{equation}
P = P-\eta\frac{\partial J}{\partial P},
\end{equation}
where $\eta$ is the learning rate. Alg.~\ref{Alg:Grad} summarizes the feed forward and the back propagation of the proposed method.

\begin{algorithm}[t]
	\caption{Attribute aware pooling method for pedestrain attribute classification with CNNs.}
	\label{Alg:Grad}
	\begin{algorithmic}[1]
		\REQUIRE{The multi-branch CNN $\mathcal{N}$, a given image $\mathbf{x}$ and its label vector $\mathbf{y}$. Two co-occurrence matrices $C$ and $\tilde{C}$, weight parameter $\lambda$, and learning rate $\eta$.}
		\STATE \textbf{Feed Forward}
		\STATE Calculate conditional probabilities from the $m$ branches and constitute $P$ (Eq.~\ref{Eq:P});
		\FOR{$l = 1$ to $m$}
		\STATE Locally max-pooling: $Q_l \leftarrow \max_{i\neq l} P_i$ (Eq.~\ref{Eq:localmax});
		\STATE Estimate $P_l^+$ using the co-occurrence priori:\\ $P_l^+ \leftarrow (Q_lC+(1-Q_l)\tilde{C})/k$ (Eq.~\ref{Eq:P+});
		\STATE Form the estimation: $\hat{P}_l \leftarrow P_l+\lambda P_l^+$ (Eq.~\ref{Eq:hatP});
		\ENDFOR
		\STATE Calculate the overall estimation $\hat{\mathbf{p}}$ of $\mathbf{x}$ (Eq.~\ref{Eq:finalest});
		\STATE \textbf{Back Propogation}
		\STATE $\nabla(\hat{\mathbf{p}})\leftarrow \hat{\mathbf{p}}-\mathbf{p}$;
		\STATE {\small$\nabla(\mathcal{E})\leftarrow \nabla(\hat{\mathbf{p}})/(\mathcal{E}^T\mathbf{1})-(\nabla(\hat{\mathbf{p}})^T\mathcal{E})/(\mathcal{E}^T\mathbf{1})^{2}$;}
		\STATE $\nabla(\hat{P}) \leftarrow M \;\text{diag}(\nabla(\mathcal{E}));$
		\STATE $\nabla(P) \leftarrow \nabla(\hat{P});$
		\FOR{$l = 1$ to $m$}
		\FOR{$j = 1$ to $k$}
		\STATE  $\nabla(P) \leftarrow \nabla(P)+$\\
		$\quad \quad \quad \frac{\lambda}{k}\nabla(\hat{P}_{l,j})S_l\;\text{diag}(C_j-\tilde{C}_j)$ (Eq.~\ref{Eq:gradP});
		\ENDFOR
		\ENDFOR
		\STATE $P\leftarrow P-\eta \nabla(P);$
		\ENSURE{The classification result $\hat{\mathbf{p}}$ and $\nabla(P)$.}
	\end{algorithmic}
\end{algorithm}

\section{Experiments}
Here we will present the setting of experiments for evaluating the proposed method and compare it with the state-of-the-art methods on benchmark pedestrian attribute recognition datasets.

\subsection{Datasets and Metrics} 
The evaluation is conducted on three largest publicly available pedestrian attribute datasets: 
\begin{itemize}
	\item PETA~\cite{deng2014pedestrian}: The PEdesTrian Attribute dataset consists of 19,000 pedestrian images with 65 attributes (61 binary and 4 multi-class). Following~\cite{li2015multi,zhao2018grouping}, we randomly divide the whole dataset into three partitions: 9,500 images for training, 1,900 for validation, and 7,600 for testing, and focus on the 35 attributes whose positive proportions are bigger than 1/20. 
	\item RAP~\cite{li2016richly}: The Richly Annotated Pedestrian attribute dataset contains 41,585 pedestrian images with 72 attributes (69 binary and 3 multi-class). It is split into 33,268 images for training and the remaining 8,317 for testing. The same 51 binary attributes are evaluated for fair comparison following~\cite{li2016richly}.
	\item PA-100K~\cite{liu2017hydraplus}: The PA-100K dataset has 100,000 images with 26 commonly used binary attributes. The whole dataset is randomly split into training, validation and test sets with a ratio of $8:1:1$.
\end{itemize}

We use five evaluation metrics including a class-centric metric mean accuracy (mA), and four instance-centric metrics, \ie accuracy, precision, recall and F1-score, following~\cite{li2016richly,liu2017hydraplus,zhao2018grouping}

\subsection{Implementation Details}
\begin{figure}[htb]
	\centering
	\includegraphics[width=0.65\linewidth]{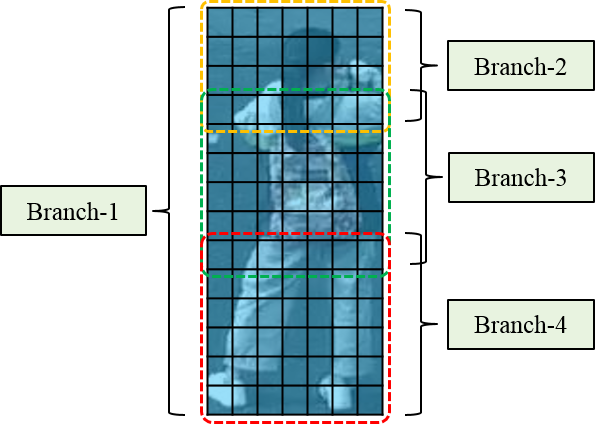}
	\caption{Feature maps parition for multi-branch architecture.}
	\label{Fig:multi-branch}
\end{figure}

In the proposed method, there are $m$ branches for attribute prediction. We use ResNet-50~\cite{ResNet} as the backbone network for building our multi-branch architecture. The main body of ResNet-50 except the last residual block is shared and the the last block is adopted as the architecutre of each branch. For a $448\times224$ input image, the feature maps $F$ from the shared CNN is of size $2048\times14\times7$. Specially, we design 4 branches in our model and the whole feature $F$ are fed into the first branch. Inspired by~\cite{yi2014deep,sun2018beyond}, we spatially partition the feature maps $F$ into 3 parts with sizes of $4\times7$, $6\times7$ and $6\times7$, serving for the other three branches, as shown in Fig~\ref{Fig:multi-branch}. Each part focuses on different human position, \ie head, upper body and lower body, for position-aware attribute recognition.

\begin{table*}[t]
	\tabcolsep3.0pt
	\begin{center}
		\small
		\renewcommand\arraystretch{1.05}
		\begin{tabular}{l|ccccc|ccccc|ccccc}
			\hline
			\multirow{2}{*}{\diagbox{Method}{Dataset}} & \multicolumn{5}{c|}{PETA} & \multicolumn{5}{c|}{RAP} & \multicolumn{5}{c}{PA-100K}\\\cline{2-16} 
			&mA& acc & prec & recall & F1 &mA& acc & prec & recall & F1 &mA& acc & prec & recall & F1  \\
			\hline
			\hline
			ACN~\cite{sudowe2015person} &81.15& 73.66& 84.06& 81.26& 82.64& 69.66& 62.61& 80.12& 72.26& 75.98& - &- &-& -& - \\
			DeepMAR~\cite{li2015multi} &82.60& 75.07& 83.68& 83.14& 83.41& 73.79& 62.02& 74.92& 76.21& 75.56& 72.7 &70.39 &82.24& 80.42& 81.32 \\
			JRL~\cite{Wang_2017_ICCV} &82.13 &-&82.55 & 82.12 & 82.02& 74.74 & - & 75.08 & 74.96& 74.62& - &- &-& -& - \\
			HP-net~\cite{liu2017hydraplus} &81.77& {76.13} & {84.92}& 83.24& 84.07& 76.12& 65.39& 77.33& 78.79& 78.05& 74.21 &72.19 &82.97& 82.09& 82.53 \\
			CTX C-RNN~\cite{Li_2017_CVPR} &80.13& -& 79.68& 80.24& 79.68& 70.13 & -&71.03& 71.20& 70.23& - &- &-& -& - \\
			SR C-RNN~\cite{Liu_2017_CVPR} &82.83& - &82.54 &82.76 &82.65& 74.21& -& 75.11 &76.52 & 75.83& - &- &-& -& - \\
			LG-Net~\cite{liu2018localization} &-&-&-&-&-& 78.68 & \underline{68.00} & \underline{80.36} & 79.82 & \underline{80.09} & \underline{79.96} & \underline{75.55} & \underline{86.99} & \underline{83.17} & \underline{85.04} \\
			VAA~\cite{sarafianos2018deep} &84.59& \underline{78.56} &\underline{86.79} & 86.12 &86.46& -& -& - &- & -& - &- &-& -& - \\
			GRL~\cite{zhao2018grouping} &\underline{86.70}& -& 84.34& \textbf{88.82} & \underline{86.51}& \underline{81.20} &-&77.70 & \textbf{80.90} & 79.29& - &- &-& -& - \\
			\hline
			Baseline & 84.68& 78.89& 85.38 &86.41 &85.83 & 79.67 & 66.01 & 77.93 & 79.25 & 78.58 & 78.12 &74.11 &84.42& 84.09& 84.25 \\
			Multi-branch & 85.60 & 79.07 & 86.05 & 87.26 & 86.65 & 80.54 & 66.84 & 79.06 & 79.64 & 79.35 & 79.26 & 76.20 &86.44& 84.08& 85.24 \\
			CoCNN &\textbf{86.97}&\textbf{79.95}&\textbf{87.58}&\underline{87.73}&\textbf{87.65}&\textbf{81.42}&\textbf{68.37}&\textbf{81.04}&\underline{80.27}&\textbf{80.65}&\textbf{80.56}&\textbf{78.30}&\textbf{89.49}&\textbf{84.36}&\textbf{86.85}\\
			\hline
		\end{tabular}
	\caption{Evaluation of CoCNN on PETA, RAP and PA-100K datasets with \textbf{bold} best result and \underline{underline} the second best result, compared with previous benchmark methods. - represents no reported result available.}
	\label{Tab:PETA}
	\end{center}
\vspace{-0.4cm}
\end{table*}

In our method, we first counted occurrence and co-occurrence numbers of different categories among train images and built two matrices $C$ and $\tilde{C}$ for each dataset. The backbone model parameters are initialized by directly using the pre-trained models on ILSVRC 2012~\cite{ImageNet}. Adam optimizer~\cite{kingma2014adam} with a batch size of 32 is used to fine-tune the entire model with an initial learning rate of 0.01. The images are resized to $512\times256$ and during training randomly cropped to the size of $448\times224$ with random horizontal flippling. All methods were implemented using PyTorch~\cite{paszke2017automatic} and run on NVIDIA V100 graphics cards. We denote the multi-branch architecture with the proposed attribute aware pooling method as CoCNN. 

The proposed CNN architecture with the attribute aware pooling method has only one hyper-parameter $\lambda$ whose functionality is to combine the original CNN output and the estimation from the context information. We tested the impact of $\lambda$ by tuning it from $0$ to $0.5$ with the step of 0.05 on the mA metric using the PETA dataset. In fact, a larger $\lambda$ makes the result $\hat{\mathbf{p}}$ more inclined to the auxiliary estimation $P^+$ and vice versa. As a result, we obtained a 86.97\% mA value at $\lambda = 0.2$ which is the best trade-off between estimations in Eq.~\ref{Eq:hatP}. Moreover, for RAP and PA-100K datasets, we keep $\lambda = 0.2$ in the following experiments.

\subsection{Results and Analysis}
\subsubsection{Effect of Multi-branch Architecture and AAP}
The improvement of CoCNN mainly comes from two aspects, \ie multi-branch architecture and attribute aware pooling. In this section, we will measure how much improvement these two aspects bring. The vanilla ResNet-50 with binary cross entropy loss is adopted as baseline model, and the multi-branch ResNet-50 without attribute aware pooling is denoted as multi-branch model. From the results in Tab.~\ref{Tab:PETA}, we can see that multi-branch architecture improve the metrics generally in the three datasets by focusing on different parts of the human body. With attribute aware pooling, CoCNN further improves the accuracy to higher level.

\begin{figure}[h]
	\centering
	\includegraphics[width=1.0\linewidth]{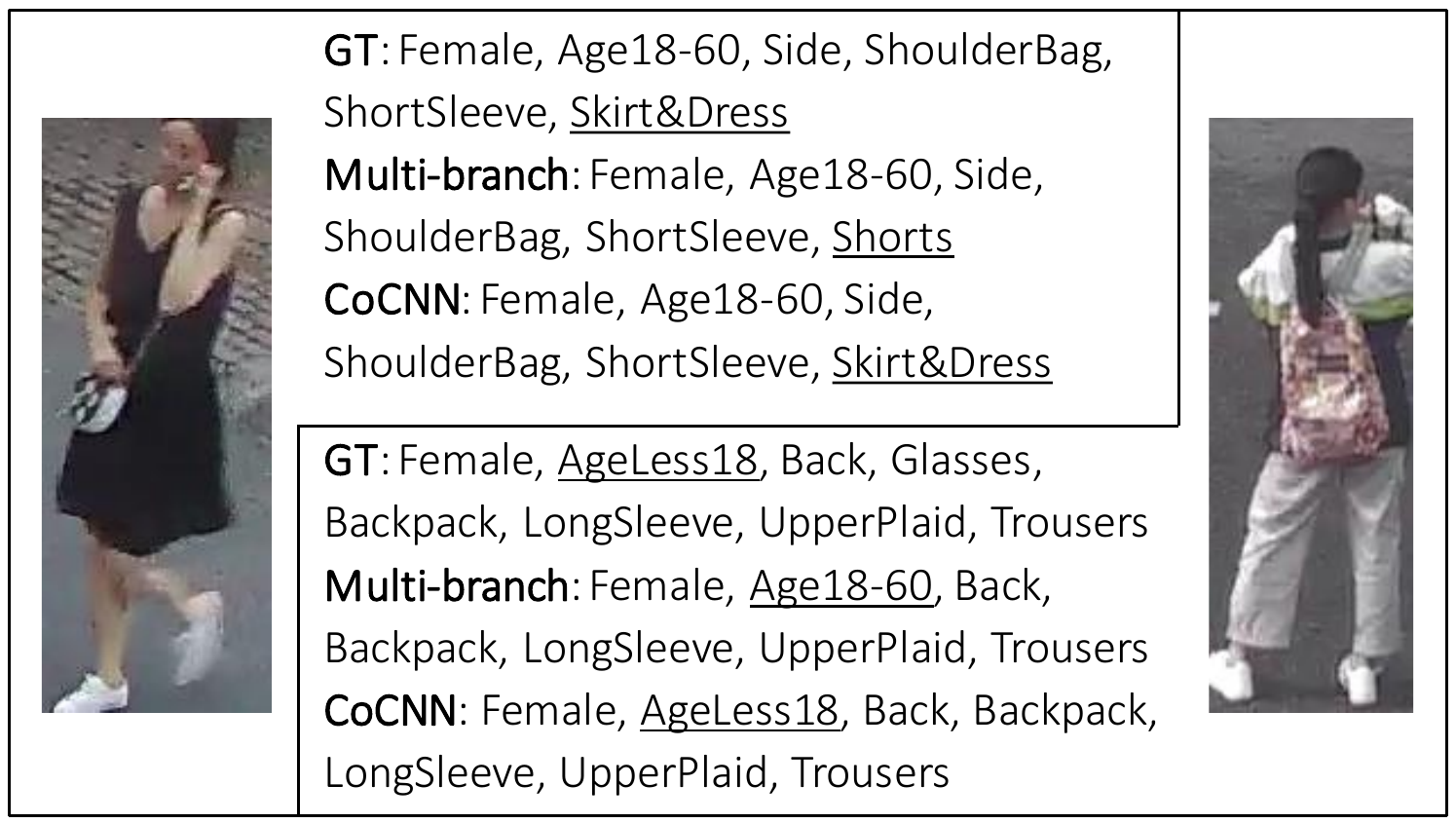}
	\vspace{-0.5cm}
	\caption{Qualitative results from PA-100K dataset of CoCNN and the multi-branch model. Wherein, GT means groundtruth, \underline{underlined} tags are attributes which are needed to be noticed.}
	\label{Fig:comp}
	\vspace{-0.5cm}
\end{figure}

Two examples from the PA-100K dataset are given in Fig.~\ref{Fig:comp} for qualitative analysis. From the results, we found that multi-branch model made wrong predictions ``Skirt\&Dress'' and ``AgeLess18'' for the two pedestrians, respectively. By exploiting their correlation with other attributes, especially the ``Female'' in the left person, and the ``UpperPlaid'' and ``Backpack'' in the right example, CoCNN well corrected the wrong predictions in multi-branch model.

\subsubsection{Comparison with State-of-the-art Methods} 
After verifing the effect of the proposed method, we compared the proposed approach with several state-of-the-art approaches, \eg DeepMAR~\cite{li2015multi}, HP-net~\cite{liu2017hydraplus}, SR C-RNN~\cite{Liu_2017_CVPR}, VAA~\cite{sarafianos2018deep}, and GRL~\cite{zhao2018grouping}. Results of comparison models are mostly reported from their papers directly. The results are listed in Tab.~\ref{Tab:PETA}.

From results on PETA dataset, we found that, CoCNN clearly outperformed other state-of-the-art methods and achieved the highest mA, acc, prec and F1 values (86.97, 79.95, 87.65). The recall of CoCNN is the second best one. This improvement mainly comes from that the proposed method takes multiple attributes into account when generating the classification annotation.

In addition, similar observation also can be seen on RAP and PA-100K datasets. This phenomenon further illustrates that the proposed attribute aware pooling method is a general auxiliary regularization for enhancing pedestrian attribute recognition and can be embedded into any similar task.

\section{Conclusions}
Existing vanilla CNNs cannot be effectively applied to handle the pedestrian attributes recognition task. Therefore, we propose using the co-occurrence priori for improving performance of CNNs, namely, attribute aware pooling. With a multi-branch CNN as base architecture, we first construct two co-occurrence tables through training sets and use context information of every branch to complete the decision derived from each branch itself, which not only excavates properties of different human parts but also investigates the relationship between different branch. Experiments conducted on several benchmark datasets show that the performance of the proposed method is better than those of state-of-the-art methods in terms of the commonly used metrics.

\clearpage
\small
\nocite{wang2018learning}
\nocite{han2018autoencoder}
\nocite{luo2018heterogeneous}
\nocite{luo2019transferring}

\bibliographystyle{named}
\bibliography{ref}

\end{document}